\PassOptionsToPackage{table}{xcolor}
\documentclass[11pt]{article}

\usepackage[preprint]{acl}
\usepackage{times}
\usepackage{latexsym}
\usepackage[T1]{fontenc}
\usepackage[utf8]{inputenc}
\usepackage{array,tabularx}
\usepackage{amsmath,amssymb,amsthm}
\usepackage{mathtools}
\usepackage{amssymb}
\usepackage{pifont}
\definecolor{bestgreen}{RGB}{222,242,222}

\usepackage{booktabs}
\usepackage{multirow}

\usepackage{algorithm}
\usepackage{algorithmic}

\usepackage{enumitem}

\usepackage{graphicx}
\usepackage{subcaption}
\usepackage{placeins}

\usepackage{url}
\usepackage{xurl}

\newtheorem{theorem}{Theorem}[section]
\newtheorem{lemma}[theorem]{Lemma}
\newtheorem{proposition}[theorem]{Proposition}

\newtheorem{definition}[theorem]{Definition}
\newtheorem{assumption}[theorem]{Assumption}

\usepackage{microtype}
\usepackage{inconsolata}
\usepackage{pifont}
\usepackage{tabularx}

\newcommand{\rwcheck}{\textcolor{green!55!black}{\ding{51}}}
\newcommand{\rwxmark}{\textcolor{red!70!black}{\ding{55}}}
\newcommand{\method}{\textsc{EnvProbe}}
\newcommand{\fieldset}{\mathcal{F}}
\newcommand{\proc}{\mathrm{proc}}
\newcommand{\spat}{\mathrm{spat}}

\title{%
	Ask the World Before Acting:\\
	Environment Probing for Calibrating Agent World Models
}

\author{\textbf{Xinyuan Song}$^{1}$ \quad
	\textbf{Zekun Cai}$^{2}$ \\
	$^{1}$Emory University, Atlanta, GA, USA \quad
	$^{2}$The University of Tokyo, Tokyo, Japan \\
	\texttt{xinyuan.song@emory.edu, caizekun@csis.u-tokyo.ac.jp} \\
}

\begin{document}
	
	\maketitle

	\begin{abstract}
		Long-horizon language agents enable autonomous decision-making in complex environments by planning and executing sequences of interdependent actions. To act reliably, these agents must maintain an accurate world model that tracks the environment's evolving state and supports subsequent decisions. However, as interactions accumulate, stored beliefs can become outdated or incorrect, causing agents to plan and act on assumptions that no longer reflect the environment. Such errors are difficult to resolve through additional reasoning or self-reflection alone, because the information needed for correction must be obtained from the environment. Actively probing the environment therefore offers a direct way to verify and correct these beliefs. To this end, we introduce \method{}, a framework that enables agents to actively explore their environments and correct inaccurate beliefs during long-horizon execution. Specifically, \method{} prioritizes belief fields for verification based on task relevance and potential errors. It then selectively probes the environment to retrieve the actual value of a chosen field and incorporates the observation into the agent's world model. With the updated beliefs, the agent revises its plan and continues execution based on the newly acquired environmental information. Extensive experiments across three controlled environments demonstrate that \method{} substantially improves terminal world-state accuracy compared with agents operating without environment probing, leading to more reliable long-horizon decision-making. These results demonstrate the importance of actively verifying and correcting beliefs through environmental interactions for maintaining accurate world models in long-horizon language agents. Our code is available at
		\url{https://github.com/Hik289/Environment-reduce-error.git}.
	\end{abstract}

	\section{Introduction}
	
	\label{sec:intro}
	
	\begin{figure*}[!ht]
		\centering
		\includegraphics[width=0.9\textwidth]{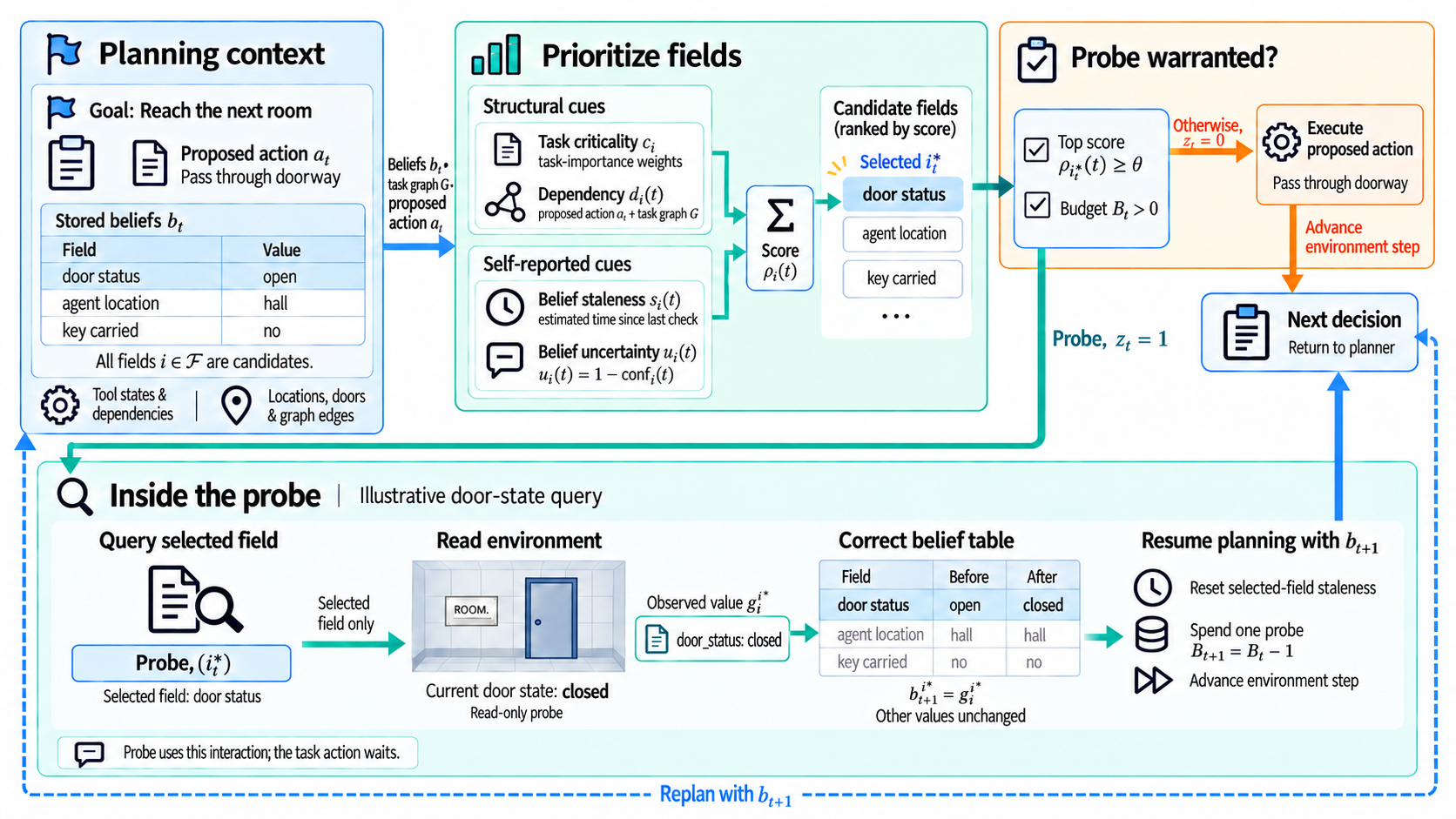}
		
		\caption{\textbf{Overview of \method{}.} (a) \textbf{Belief prioritization:} The agent ranks belief fields based on task relevance, dependencies, staleness, and uncertainty. (b) \textbf{Environment probing:} The agent selects what to probe in the environment based on field priority and the available budget. (c) \textbf{Environment belief correction:} The agent retrieves the environment state, updates the corresponding belief, and replans using the corrected world model.}
		\label{fig:intuition}
		
	\end{figure*}
	
	Long-horizon language agents enable autonomous decision-making in complex environments by planning and executing sequences of interdependent actions. Building on reasoning-and-acting frameworks such as ReAct~\citep{yao2023react}, reflection-based agents such as Reflexion~\citep{shinn2023reflexion}, and search-based planning methods such as LATS~\citep{zhou2024lats}, these agents increasingly support complex interactions with external environments. Their capabilities extend from tool use~\citep{schick2023toolformer,qin2024toolllm} to embodied decision-making~\citep{huang2022inner,ahn2022saycan} and long-horizon tasks evaluated in ALFWorld~\citep{shridhar2021alfworld}, ScienceWorld~\citep{wang2022scienceworld}, WebArena~\citep{zhou2024webarena}, and UltraHorizon~\citep{luo2025ultrahorizon}. However, reliable long-horizon execution requires more than generating appropriate actions at individual steps, as effective planning depends on anticipating future environment states~\citep{zhang2026internalizing}. Agents must maintain an accurate world model that represents the evolving environment and supports subsequent planning decisions, as demonstrated by Reasoning via Planning (RAP)~\citep{hao2023rap}. Such a model allows agents to track relevant environmental states, reason about action preconditions, and coordinate dependencies across multiple stages of a task~\citep{hill2025language}.
	
	A fixed world model becomes increasingly unreliable during long-horizon execution as environmental changes render stored beliefs outdated. In interactive environments such as ALFWorld~\citep{shridhar2021alfworld} and WebArena~\citep{zhou2024webarena}, agents may continue relying on states that no longer reflect the environment, leading to discrepancies between their internal beliefs and actual conditions~\citep{yin2026glove}. For example, an agent may assume that a tool remains available after its state has changed or attempt to traverse a route that is no longer accessible. Such belief errors can propagate through subsequent decisions, undermining long-horizon execution even when individual reasoning steps are logically consistent~\citep{wang2026whyfails,luo2025ultrahorizon}. Although reflection-based approaches such as Reflexion~\citep{shinn2023reflexion} and Self-Refine~\citep{madaan2023selfrefine} improve decisions through feedback and iterative reasoning, they cannot reliably correct beliefs without access to the missing environmental evidence.
	
	This limitation motivates actively acquiring information from the environment to verify and correct the agent's beliefs. Rather than relying exclusively on previously observed states, an agent can interact with its environment to determine whether a relevant belief remains valid. For instance, it may inspect the current state of a tool before invoking a dependent operation or revisit a location before planning its next movement. Such interactions provide direct evidence for updating the world model and enable subsequent decisions to reflect the environment's actual state. However, environmental verification introduces an additional decision problem: agents must determine which beliefs require verification and when such verification should occur. Because each probe consumes an interaction step, unnecessary verification can reduce the opportunities available for task execution. This creates a fundamental trade-off between acquiring information to correct the world model and executing actions to advance the task, closely related to partially observable planning~\citep{kaelbling1998planning}, online information gathering~\citep{ross2008online}, and Bayesian experimental design~\citep{chaloner1995bayesian}.
	
	To address this challenge, we introduce \method{}, a framework that enables language agents to actively probe their environments and correct inaccurate beliefs during long-horizon execution. \method{} integrates environment probing into the planning process to obtain direct evidence about the current environment state. Specifically, \method{} evaluates belief fields based on their task relevance and potential for error, then selectively queries the environment to verify the most relevant field under a interaction budget. The acquired evidence is used to correct the corresponding belief in the agent's world model, enabling the agent to replan and continue execution with updated environmental information. Figure~\ref{fig:intuition} illustrates the overall workflow of \method{}.
	
	We provide a type-stratified theoretical analysis of how environment probing improves belief accuracy while consuming task-action opportunities, highlighting differences between procedural and spatial beliefs. We further evaluate \method{} across three controlled interactive environments covering procedural tool dependencies and spatial reasoning tasks. Extensive experiments demonstrate that \method{} substantially improves terminal world-state accuracy compared with agents operating without environment probing. Additional analyses examine the effects of probing frequency, belief selection, and verification signals, demonstrating the effectiveness of actively acquiring environmental evidence for reliable world-model maintenance during long-horizon execution.
	
	Our contributions are summarized as follows:
	\begin{itemize}[left = 0em]
		\item We identify the challenge of correcting outdated world models through environment probing, highlighting the trade-off between acquiring environmental evidence and advancing task execution in long-horizon agents.
		\item We propose \method{}, an environment-probing framework that selectively interacts with the environment to verify beliefs, correct inaccurate world models, and support replanning during long-horizon execution.
		\item We develop a theoretical analysis that characterizes the benefits of environment probing for belief correction and its interaction costs, revealing distinct effects on procedural and spatial beliefs.
		\item We conduct extensive experiments across three controlled environments, demonstrating that \method{} improves terminal world-state accuracy and investigating the effects of different probing strategies and verification signals.
	\end{itemize}

	\begin{table*}[!ht]
		\centering
		\small
		\setlength{\tabcolsep}{4pt}
		\renewcommand{\arraystretch}{1.2}
		\caption{\textbf{Comparison with related work.} We compare existing approaches in terms of environment probing, belief correction, task-aware selection, probing cost, and belief-type analysis. \rwcheck\ indicates support for a property, while \rwxmark\ indicates its absence.}
		
		\label{tab:related_novelty}
		\begin{tabularx}{\textwidth}{@{}l*{5}{>{\centering\arraybackslash}X}@{}}
			\toprule
			\textbf{Method}
			& \textbf{Environment Probing}
			& \textbf{Belief Correction}
			& \textbf{Task-Aware Selection}
			& \textbf{Probing Cost}
			& \textbf{Belief-Type Analysis} \\
			\midrule
			BB-WM~\citep{kumar2026belief}
			& \rwxmark & \rwcheck & \rwxmark & \rwxmark & \rwxmark \\
			GLOVE~\citep{yin2026glove}
			& \rwcheck & \rwcheck & \rwxmark & \rwcheck & \rwxmark \\
			RPMS~\citep{yuan2026rpms}
			& \rwxmark & \rwcheck & \rwxmark & \rwxmark & \rwxmark \\
			InfoSeeker~\citep{fang2025infoseeker}
			& \rwcheck & \rwxmark & \rwcheck & \rwcheck & \rwxmark \\
			BKLVA~\citep{zhao2025seeing}
			& \rwcheck & \rwcheck & \rwcheck & \rwxmark & \rwxmark \\
			ReSpAct~\citep{dongre2024respact}
			& \rwxmark & \rwxmark & \rwcheck & \rwxmark & \rwxmark \\
			UoT~\citep{hu2024uot}
			& \rwxmark & \rwxmark & \rwcheck & \rwxmark & \rwxmark \\
			\midrule
			\rowcolor{blue!9}
			\textbf{\method{} (Ours)}
			& \rwcheck & \rwcheck & \rwcheck & \rwcheck & \rwcheck \\
			\bottomrule
		\end{tabularx}
		
	\end{table*}

	\section{Related Work}
	
	\label{sec:related}
	
	\paragraph{Language agents and world models.}
	Long-horizon language agents combine reasoning, planning, and environment interaction to accomplish complex tasks. ReAct~\citep{yao2023react} interleaves reasoning with actions, Reflexion~\citep{shinn2023reflexion} incorporates feedback, and LATS~\citep{zhou2024lats} introduces search-based planning. Toolformer~\citep{schick2023toolformer} and ToolLLM~\citep{qin2024toolllm} enable external tool use, while embodied agents~\citep{huang2022inner,ahn2022saycan} and Voyager~\citep{wang2024voyager} support environment interaction and persistent memory. Recent work on memory-environment realignment~\citep{yin2026glove} and rule-augmented memory~\citep{yuan2026rpms} further addresses belief consistency. However, these approaches do not explicitly study budgeted environment probing for correcting individual belief fields during long-horizon execution.
	
	\paragraph{Active information gathering and partial observability.}
	Partially observable Markov decision processes (POMDPs)~\citep{kaelbling1998planning} and online planning~\citep{ross2008online} formalize decision-making under incomplete observations, while Bayesian experimental design~\citep{chaloner1995bayesian} and adaptive submodularity~\citep{golovin2011adaptive} study information acquisition under limited resources. Unlike general information-gathering methods, \method{} focuses on selectively querying the environment to correct structured beliefs maintained by language agents, explicitly accounting for the interaction cost of verification.
	
	\paragraph{LLM uncertainty and information seeking.}
	Confidence calibration has been studied in neural networks~\citep{guo2017calibration} and language models~\citep{kadavath2022language}. UoT~\citep{hu2024uot} uses uncertainty to guide question generation, InfoSeeker~\citep{fang2025infoseeker} explores information gathering under partial observability, and ReSpAct~\citep{dongre2024respact} incorporates clarification actions. However, model-reported uncertainty may be unreliable when incorrect beliefs are held with high confidence. \method{} combines task structure with belief-level signals to guide environment probing and analyzes the limitations of uncertainty-only verification.
	
	\paragraph{Long-horizon agent evaluation.}
	Benchmarks such as ALFWorld~\citep{shridhar2021alfworld}, WebArena~\citep{zhou2024webarena}, VisualWebArena~\citep{koh2024visualwebarena}, Mind2Web~\citep{deng2023mind2web}, AgentBench~\citep{liu2023agentbench}, and UltraHorizon~\citep{luo2025ultrahorizon} primarily evaluate task completion, making it difficult to isolate failures caused by inaccurate world models. Our controlled environments provide ground-truth belief states, enabling direct evaluation of terminal world-state accuracy, useful-probe rates, and belief-error dynamics.
	
	\paragraph{Comparison to Prior work.}
	Table~\ref{tab:related_novelty} compares \method{} with representative approaches to belief estimation~\citep{kumar2026belief}, memory correction~\citep{yin2026glove,yuan2026rpms}, active information gathering~\citep{fang2025infoseeker,zhao2025seeing}, and clarification-based planning~\citep{dongre2024respact,hu2024uot}. Unlike prior approaches, \method{} jointly considers environment probing, field-level belief correction, task-aware selection, and probing costs, with a type-stratified analysis of procedural and spatial beliefs.

	\section{Problem Setup}
	\label{sec:prelim}
	
	We consider a long-horizon language agent that maintains a structured world model to support planning and interaction with a dynamic environment. The agent can either execute task actions or probe the environment to verify and correct its stored beliefs.
	
	\paragraph{Belief world model.}
	Let $\fieldset=\{1,\ldots,n\}$ denote the set of belief fields. At step $t$, each field $i\in\fieldset$ has a stored belief $b_t^i$ and a corresponding ground-truth environment value $g_t^i$, both belonging to $\mathcal{V}_i$. The agent's world model is $b_t=(b_t^1,\ldots,b_t^n)$, and its terminal accuracy is measured by
	\begin{equation}
	A_H=\frac{1}{n}\sum_{i=1}^{n}\mathbf{1}\{b_H^i=g_H^i\},
	\label{eq:terminal_accuracy}
	\end{equation}
	where $H$ is the episode horizon. World-model accuracy is distinct from task success, as additional verification may reduce the number of available task actions.

	\paragraph{Environment probing.}
	At step $t$, the agent may probe a belief field $i$ to retrieve its current environment value and update the corresponding belief:
	\begin{equation}
	\textsc{Probe}_t(i):
	b_{t+1}^i=g_t^i,\quad
	b_{t+1}^j=b_t^j\quad(j\neq i).
	\label{eq:probe_update}
	\end{equation}
	Each probe consumes one interaction step without executing a task action. Given a horizon $H$, the number of probes is constrained by a budget $B$:
	\begin{equation}
	\sum_{t=0}^{H-1}\mathbf{1}\{a_t\in\{\textsc{Probe}(i):i\in\fieldset\}\}\leq B.
	\label{eq:probe_budget}
	\end{equation}
	
	\paragraph{Belief types.}
	We distinguish two types of belief fields based on how environmental changes affect them. The first type captures action-dependent states, such as tool availability, action dependencies, and subgoal completion, which may change as the agent executes tasks. The second type captures spatial configurations, such as object locations, door states, and graph connectivity, which may change independently of the agent's actions. This distinction matters because action history can help identify the former changes but may provide limited evidence about the latter.
	
	\paragraph{Probe policies.}
	At each interaction step, the agent decides whether to execute a task action or probe the environment to verify a belief. We consider seven policies: \textbf{No-Probe}, which executes without explicit verification; \textbf{Random-Probe} and \textbf{Periodic-Probe}, which perform non-adaptive probing; \textbf{Self-Uncertainty}, which selects probes based on model-reported uncertainty, following uncertainty sampling~\citep{lewis1994sequential,settles2009active} and language-model confidence estimation~\citep{kadavath2022language}; \textbf{\method{}-Simple} and \textbf{\method{}-Judge}, which use task-aware belief selection with different verification signals; and \textbf{Oracle-Probe}, which uses ground-truth belief errors as an upper-bound diagnostic. Figure~\ref{fig:intuition} illustrates how environment probing supports belief correction and subsequent replanning.

	\section{\method{}: Environment Evidence During Planning}
	\label{sec:method}
	
	\method{} enables language agents to correct outdated world models through selective environment probing during long-horizon execution. Before executing a planned action, the agent evaluates belief fields using task relevance and reliability signals to determine whether verification is needed. If probing is selected, the observed environment value corrects the corresponding belief, and the agent replans accordingly. Otherwise, it executes the planned action. This process balances belief accuracy against probing costs without modifying the underlying planner, as illustrated in Figure~\ref{fig:intuition}.

	\subsection{Belief Field Scoring}
	\label{sec:method:score}
	
	At each step $t$, \method{} evaluates each belief field $i\in\fieldset$ using four signals that capture its task relevance and potential for error:
	
	\begin{itemize}[left=0em]
		\item \textbf{Task criticality.} Measures the importance of field $i$ to task completion using its task-importance weight $w_i$:
		\begin{equation}\label{eq:criticality}c_i=\frac{w_i}{\max_{j\in\fieldset}w_j}.\end{equation}
		
		\item \textbf{Dependency relevance.} Measures the relationship between field $i$ and the proposed action $a_t$ in the task graph $G$:
		\begin{equation}\label{eq:dependency}d_i(t)=D(i,a_t,G)\in[0,1].\end{equation}
		Direct dependencies receive higher scores than indirect or unrelated fields.
		
		\item \textbf{Belief staleness.} Measures the estimated time $\hat{\tau}_i(t)$ since field $i$ was last verified:
		\begin{equation}\label{eq:staleness}s_i(t)=S(\hat{\tau}_i(t))\in[0,1],\end{equation}
		where $S$ is a nondecreasing normalization function.
		
		\item \textbf{Belief uncertainty.} Measures uncertainty using the agent's reported confidence in field $i$:
		\begin{equation}\label{eq:uncertainty}u_i(t)=1-\mathrm{conf}_i(t).\end{equation}
	\end{itemize}
	
	\paragraph{Unified scoring.}
	Using the four signals defined in Eqs.~\eqref{eq:criticality}--\eqref{eq:uncertainty}, \method{} assigns each field a verification priority:
	\begin{equation}\label{eq:rho_def}\rho_i(t)=c_i+d_i(t)+s_i(t)+u_i(t).\end{equation}
	The score separates task-related structural information from the agent's self-reported belief reliability:
	\begin{equation}\label{eq:score_decomposition}\rho_i(t)=\underbrace{c_i+d_i(t)}_{\rho_i^{\mathrm{str}}(t)}+\underbrace{s_i(t)+u_i(t)}_{\rho_i^{\mathrm{self}}(t)}.\end{equation}
	As shown in Eq.~\eqref{eq:score_decomposition}, the structural component captures task criticality and action dependencies, while the self-reported component captures estimated staleness and uncertainty. This distinction allows us to evaluate whether self-reported signals improve probe selection beyond structural information alone.

	\subsection{Probe Selection and Belief Correction}
	\label{sec:method:selection}
	
	\paragraph{Probe selection.}
	Given the verification scores in Eq.~\eqref{eq:rho_def}, \method{} selects the highest-priority belief field:
	\begin{equation}\label{eq:probe_selection}i_t^\star=\arg\max_{i\in\fieldset}\rho_i(t).\end{equation}
	The agent probes this field only if its score exceeds a threshold $\theta$ and the remaining probe budget $B_t$ is positive:
	\begin{equation}\label{eq:probe_decision}z_t=\mathbf{1}\{\rho_{i_t^\star}(t)\geq\theta\land B_t>0\}.\end{equation}
	Here, $z_t=1$ indicates probing, while $z_t=0$ indicates execution of the proposed task action. This decision balances the priority of belief verification against the limited interaction budget defined in Eq.~\eqref{eq:probe_budget}.
	
	\paragraph{Belief correction and replanning.}
	When $z_t=1$, the agent observes the environment value $g_t^{i_t^\star}$ and corrects the selected belief according to Eq.~\eqref{eq:probe_update}, leaving other fields unchanged. It then resets the field's estimated staleness, decreases the remaining probe budget, and replans using the corrected world model. When $z_t=0$, the agent executes its proposed task action and updates its beliefs using execution feedback. Each probe consumes one interaction step without advancing the task directly, creating a trade-off between improving belief accuracy and preserving task-action opportunities.

	\subsection{Algorithm and Variants}
	\label{sec:method:algo}
	
	\begin{algorithm}[!ht]
		\caption{\method{}: Environment Probing and Belief Correction}
		\label{alg:simple}
		\begin{algorithmic}[1]
			\REQUIRE Initial beliefs $\mathbf{b}_0$, task graph $G$, horizon $H$, probe budget $B$, threshold $\theta$
			\STATE $B_0\leftarrow B$
			\FOR{$t=0,\ldots,H-1$}
			\STATE Propose task action $a_t$ based on $\mathbf{b}_t$
			\FOR{each field $i\in\fieldset$}
			\STATE Compute $c_i$, $d_i(t)$, $s_i(t)$, and $u_i(t)$
			\STATE Compute $\rho_i(t)$ using Eq.~\eqref{eq:rho_def}
			\ENDFOR
			\STATE Select $i_t^\star$ using Eq.~\eqref{eq:probe_selection}
			\STATE Compute $z_t$ using Eq.~\eqref{eq:probe_decision}
			\IF{$z_t=1$}
			\STATE Observe environment value $g_t^{i_t^\star}$
			\STATE Update beliefs using Eq.~\eqref{eq:probe_update}
			\STATE $\hat{\tau}_{i_t^\star}(t+1)\leftarrow 0$
			\STATE $B_{t+1}\leftarrow B_t-1$
			\STATE Replan using corrected beliefs $\mathbf{b}_{t+1}$
			\ELSE
			\STATE Execute proposed task action $a_t$
			\STATE Update $\mathbf{b}_{t+1}$ from execution feedback
			\STATE $B_{t+1}\leftarrow B_t$
			\ENDIF
			\ENDFOR
			\RETURN Final beliefs $\mathbf{b}_H$
		\end{algorithmic}
	\end{algorithm}
	
	\paragraph{\method{}-Simple.}
	Algorithm~\ref{alg:simple} implements the threshold-based policy in Eq.~\eqref{eq:probe_decision}. The agent selects the highest-scoring field using Eq.~\eqref{eq:probe_selection} and probes it when the threshold and budget constraints are satisfied. The observed value corrects the corresponding belief according to Eq.~\eqref{eq:probe_update}, followed by replanning; otherwise, the agent executes the proposed action.
	
	\paragraph{\method{}-Judge.}
	This variant introduces a secondary language model to evaluate the selected probe, following reflection and agent-critique approaches~\citep{shinn2023reflexion,zhou2024lats}. Given the current beliefs, proposed action, and highest-scoring candidates, the judge may override the decision in Eq.~\eqref{eq:probe_decision}.
	
	\paragraph{Structural variant.}
	To isolate structural information, \textbf{\method{}-$(c+d)$}\label{def:cd_variant} removes staleness and uncertainty from Eq.~\eqref{eq:rho_def}:
	\begin{equation}\label{eq:structural_variant}\rho_i^{(c+d)}(t)=c_i+d_i(t).\end{equation}
	This variant retains the same probe selection and belief correction procedure while relying only on task criticality and dependency relevance.
	
	\paragraph{Oracle variants.}
	\textbf{Oracle-Probe} selects fields with known belief errors, while \textbf{Oracle-TW}\label{def:oracle_tw} additionally weights errors by task importance:
	\begin{equation}\label{eq:oracle_tw}i_t^{\mathrm{oracle}}=\arg\max_{i\in\fieldset}w_i\mathbf{1}\{b_t^i\neq g_t^i\}.\end{equation}
	Both serve as diagnostic references because they require ground-truth information unavailable to practical agents.

	\section{Experiments}
	\label{sec:experiments}

	\subsection{Experimental Setup}
	\label{sec:setup}
	
	\paragraph{Environments.}
	We evaluate three controlled environments with observable ground-truth states, enabling direct measurement of belief drift. \textbf{ObjectStateWorld} models mutable object locations and lock states, resembling embodied tasks such as ALFWorld~\citep{shridhar2021alfworld} and ScienceWorld~\citep{wang2022scienceworld}. \textbf{ToolDAGWorld} captures procedural dependencies and changing tool states, motivated by Toolformer~\citep{schick2023toolformer} and ToolLLM~\citep{qin2024toolllm}. \textbf{GraphNavWorld} models partially observable navigation with dynamic graph connectivity, complementing navigation and interaction benchmarks such as WebArena~\citep{zhou2024webarena}, Mind2Web~\citep{deng2023mind2web}, AgentBench~\citep{liu2023agentbench}, and UltraHorizon~\citep{luo2025ultrahorizon}. Each probe reveals the ground-truth value of a selected field at the cost of one interaction step.
	
	\paragraph{Experimental settings.}
	We evaluate low-, medium-, and high-mutation regimes, using the medium regime for primary comparisons and the others for robustness analysis. All methods share the same interaction horizon, probe budget, and initial conditions, with paired seeds ensuring controlled comparisons.
	
	\paragraph{Baselines.}
	We compare \method-Simple and \method-Judge against No-Probe and Periodic-Probe, motivated by partially observable planning~\citep{kaelbling1998planning,ross2008online}; Random-Probe; Self-Uncertainty, based on uncertainty sampling~\citep{lewis1994sequential,settles2009active} and LLM information seeking~\citep{hu2024uot,fang2025infoseeker}; and Oracle-Probe. \method-Judge incorporates an LLM critic inspired by reflection-based agents~\citep{shinn2023reflexion,zhou2024lats}. Oracle-Probe and Oracle-TW serve as privileged diagnostic references.
	
	\paragraph{Metrics and statistics.}
	Our primary metric is terminal world-state accuracy (WSA, $A_H$), defined in Eq.~\eqref{eq:terminal_accuracy}. Secondary metrics include task success (TS), useful-probe rate (UPR), and collapse onset ($\tau_c$). We use paired evaluations with bootstrap confidence intervals~\citep{efron1994introduction} and multiple-comparison correction.
	
	\paragraph{Agent implementation.}
	All methods use GPT-4o mini~\citep{openai2024gpt4omini} with deterministic decoding. The agent updates field-level beliefs, staleness estimates, and confidence scores after each interaction. \method-Judge uses the same backbone for its critic, while \method-Simple applies the score in Eq.~\eqref{eq:rho_def} directly. Staleness is normalized as $s_i(t)=\min(1,\hat\tau_i/10)$.

	\subsection{Results}
	\label{sec:results}
	
	\subsubsection{Result: Probing Repairs the World Model}
	\label{sec:results:g1}
	\begin{table*}[!ht]
		\centering
		\caption{\textbf{Terminal world-state accuracy under environment probing.} The upper block compares \method-Simple with Periodic-Probe; the lower block compares score ablations with the full \method-Simple score on ToolDAGWorld. $\hat\Delta$ denotes the absolute accuracy difference (\%).}
		
		\label{tab:main}
		\small
		\setlength{\tabcolsep}{6pt}
		\begin{tabular}{lrrrr}
			\toprule
			\textbf{Environment / Comparison} & $n$ & \textbf{Method $A_H$} & \textbf{Reference $A_H$} & $\hat\Delta$ (\%) \\
			\midrule
			Procedural (ToolDAGWorld) & 220 & \cellcolor{bestgreen}\textbf{0.431} & 0.313 & +11.76 \\
			Combined (all environments) & 660 & \cellcolor{bestgreen}\textbf{0.371} & 0.306 & +6.45 \\
			Spatial (GraphNavWorld + ObjectStateWorld) & 440 & \cellcolor{bestgreen}\textbf{0.341} & 0.303 & +3.79 \\
			\midrule
			\multicolumn{5}{l}{\textit{Score ablations on ToolDAGWorld; reference: full \method-Simple}} \\
			Structural-only vs.\ full score & 220 & \cellcolor{bestgreen}\textbf{0.491} & 0.431 & +6.03 \\
			Without uncertainty vs.\ full score & 220 & \cellcolor{bestgreen}\textbf{0.488} & 0.431 & +5.76 \\
			\bottomrule
		\end{tabular}
		
	\end{table*}
	
	\paragraph{Belief repair across environments.}
	Table~\ref{tab:main} shows that \method-Simple consistently improves terminal world-state accuracy over Periodic-Probe, with gains of +11.76\% on procedural tasks and +3.79\% on spatial tasks. The smaller spatial improvement is consistent with Lemma~\ref{lem:decomp_rev}, suggesting that self-reported reliability provides weaker guidance when environmental changes are externally induced.
	
	\paragraph{Selection signals and task trade-off.}
	The ablations in Table~\ref{tab:main} compare the full score in Eq.~\eqref{eq:rho_def} with Structural-only (Eq.~\eqref{eq:structural_variant}) and Without uncertainty, which removes $u_i(t)$. Both improve belief accuracy, but removing uncertainty substantially reduces task success (8.2\% versus 20.5\%). This demonstrates that maximizing belief repair alone can compromise task completion, consistent with Theorem~\ref{thm:tradeoff}.
	
	\begin{figure}[!ht]
		\centering
		\includegraphics[width=\columnwidth]{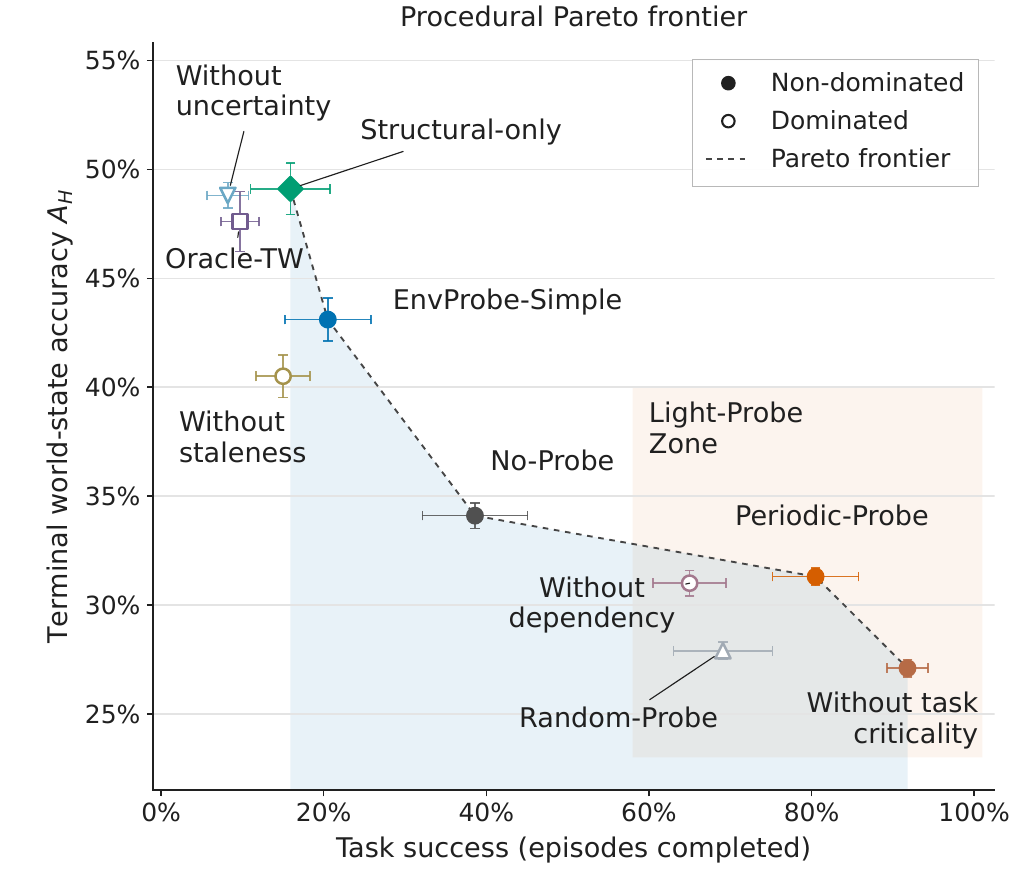}
		
		\caption{\textbf{Empirical belief--task Pareto frontier on ToolDAGWorld.} Task success versus terminal accuracy $A_H$. Filled markers indicate nondominated policies; hollow markers indicate dominated policies. Probe-intensive methods favor belief accuracy, while lighter probing favors task completion.}
		
		\label{fig:pareto_proc}
		
	\end{figure}
	
	Figure~\ref{fig:pareto_proc} illustrates the competing effects of belief repair and task execution. Additional probing can improve accuracy while consuming steps needed for task completion, consistent with the belief-gain bound in Eq.~\eqref{eq:frontier_belief} and the action-displacement bound in Eq.~\eqref{eq:frontier_task}.

	\subsubsection{ Targeted Probing vs. Probe Frequency}
	\label{sec:results:g5}
	
	\begin{table}[!ht]
		\centering
		\caption{\textbf{Probing-policy comparisons under medium stress.} Terminal world-state accuracy $A_H$.}
		
		\label{tab:ordering}
		\small
		\setlength{\tabcolsep}{4pt}
		\resizebox{\columnwidth}{!}{
			\begin{tabular}{lrr}
				\toprule
				\textbf{Comparison} & \textbf{Policy $A_H$} & \textbf{Reference $A_H$} \\
				\midrule
				No-Probe vs.\ Self-Uncertainty (spatial) & \cellcolor{bestgreen}\textbf{0.387} & 0.352 \\
				Self-Uncertainty vs.\ Periodic-Probe (spatial) & 0.303 & \cellcolor{bestgreen}\textbf{0.307} \\
				\method-Simple vs.\ Periodic-Probe (procedural) & \cellcolor{bestgreen}\textbf{0.431} & 0.313 \\
				\method-Judge vs.\ \method-Simple (spatial) & \cellcolor{bestgreen}\textbf{0.341} & 0.339 \\
				\method-Judge vs.\ Oracle-Probe & 0.341 & \cellcolor{bestgreen}\textbf{0.364} \\
				\bottomrule
			\end{tabular}
		}
		
	\end{table}
	
	Table~\ref{tab:ordering} shows that targeted probing matters more than probe frequency. On procedural tasks, \method-Simple outperforms Periodic-Probe by 11.76\%, highlighting the importance of structural cues. \method-Judge offers modest additional gains, while uncertainty-based probing remains vulnerable to confident errors. Oracle comparisons further suggest that correcting more beliefs does not necessarily improve task-relevant accuracy.
	
	\subsubsection{When and Why Probing Works}
	\label{sec:results:secondary}
	
\begin{figure*}[!ht]
		\centering
		\includegraphics[width=0.9\textwidth]{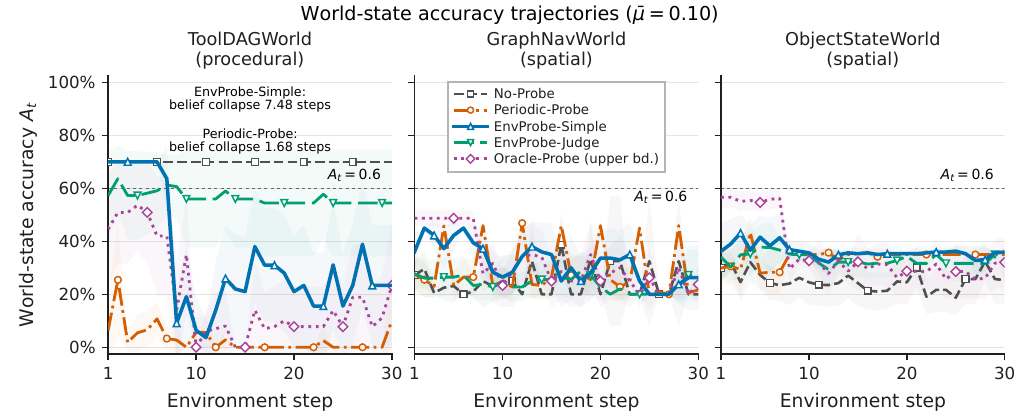}
		
		\caption{\textbf{World-state accuracy under medium stress.} \method-Simple delays belief collapse ($A_t<0.6$) on ToolDAGWorld, while spatial environments exhibit saturated collapse onset due to low initial accuracy.}
		\label{fig:collapse_curves}
		
	\end{figure*}

	\paragraph{Collapse-onset delay.}
	On ToolDAGWorld, \method-Simple delays belief collapse from 1.68 to 7.48 steps relative to Periodic-Probe. As shown in Figure~\ref{fig:collapse_curves}, spatial environments often begin near or below the accuracy threshold, making collapse onset less informative for spatial evaluation.
	
\begin{figure}[!ht]
		\centering
		\includegraphics[width=\columnwidth]{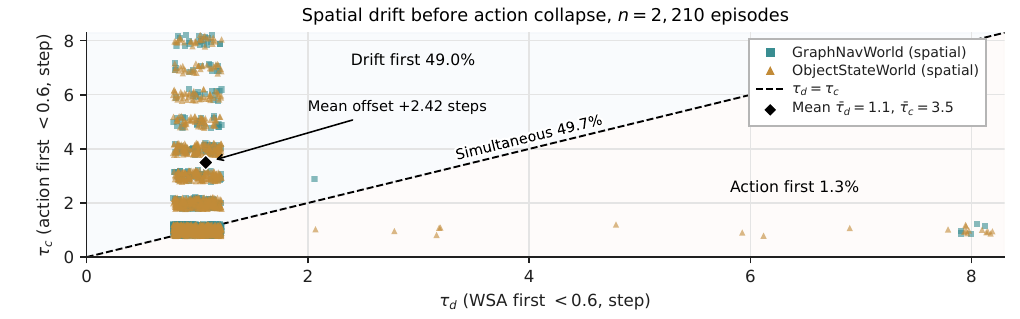}
		
		\caption{\textbf{Belief drift precedes action collapse in spatial environments.} The scatter compares belief drift ($\tau_d$) and action collapse ($\tau_c$) at the 0.6 threshold. Points above the diagonal indicate drift-first episodes.}
		\label{fig:drift_appendix}
		
	\end{figure}

	\paragraph{Drift before action collapse.}
	Figure~\ref{fig:drift_appendix} and Table~\ref{tab:drift_timing} show that belief drift precedes action collapse by 2.42 steps on average in spatial environments, whereas procedural tasks exhibit more frequent action-first failures. This difference reflects the stronger coupling between tool-precondition beliefs and action validity: a single incorrect prerequisite can invalidate an action before aggregate belief accuracy collapses.
	
	\begin{table}[!ht]
		\centering
		\caption{\textbf{Belief drift and action-collapse timing.} Positive offsets indicate that belief drift occurs before action collapse.}
		\label{tab:drift_timing}
		\small
		\setlength{\tabcolsep}{4pt}
		\resizebox{\columnwidth}{!}{
			\begin{tabular}{lrrrrr}
				\toprule
				\textbf{Environment} & $n$ & \textbf{Mean} & \textbf{Median} & \textbf{Drift-first} & \textbf{Action-first} \\
				\midrule
				Combined & 2813 & +2.22 & 0.0 & 49.2\% & 11.8\% \\
				Spatial & 2210 & \cellcolor{bestgreen}\textbf{+2.42} & 0.0 & 49.0\% & 1.3\% \\
				Procedural & 603 & +1.45 & $-2.0$ & 49.6\% & 50.4\% \\
				\bottomrule
			\end{tabular}
		}
		
	\end{table}

\begin{figure}[!ht]
		\centering
		\includegraphics[width=\columnwidth]{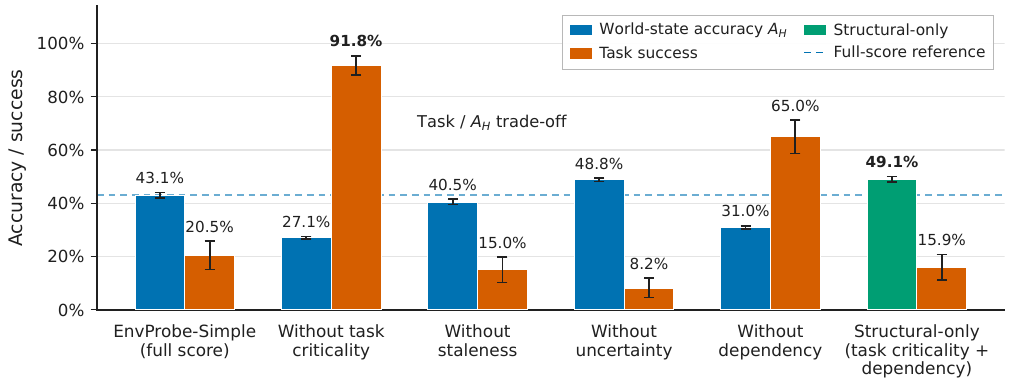}
		
		\caption{\textbf{Component ablation: belief accuracy and task success.} Blue/teal bars show $A_H$; amber bars show task success. Structural cues are essential for belief accuracy, while removing uncertainty improves $A_H$ at the expense of task success.}
		\label{fig:ablation}
		
	\end{figure}

	\paragraph{Component ablation.}
	\label{sec:results:ablation}
	Table~\ref{tab:ablation} and Figure~\ref{fig:ablation} examine the contributions of the four signals in Eq.~\eqref{eq:rho_def}. Removing task criticality or dependency substantially reduces belief accuracy, confirming the importance of structural information. Staleness provides a smaller benefit, whereas removing uncertainty improves belief accuracy but reduces task success. The structural-only score in Eq.~\eqref{eq:structural_variant} achieves the highest observed procedural accuracy, consistent with the limited value of self-reported reliability characterized by Lemma~\ref{lem:decomp_rev}.
	
	\begin{table}[!ht]
		\centering
		\caption{\textbf{Component ablation on ToolDAGWorld.} Accuracy changes (\%) relative to full \method-Simple; green highlights the best result.}
		
		\label{tab:ablation}
		\small
		\setlength{\tabcolsep}{8pt}
		\begin{tabular}{lrr}
			\toprule
			\textbf{Method / Ablation} & \textbf{$A_H$} & $\hat\Delta A_H$ (\%) \\
			\midrule
			Full \method-Simple & 0.431 & --- \\
			Without task criticality & 0.271 & $-15.92$ \\
			Without dependency & 0.310 & $-12.08$ \\
			Without staleness & 0.405 & $-2.59$ \\
			Without uncertainty & 0.488 & $+5.76$ \\
			\rowcolor{bestgreen}
			\textbf{Structural-only} & \textbf{0.491} & $\boldsymbol{+6.03}$ \\
			\bottomrule
		\end{tabular}
		
	\end{table}
	
	\begin{figure}[!ht]
		\centering
		\includegraphics[width=\columnwidth]{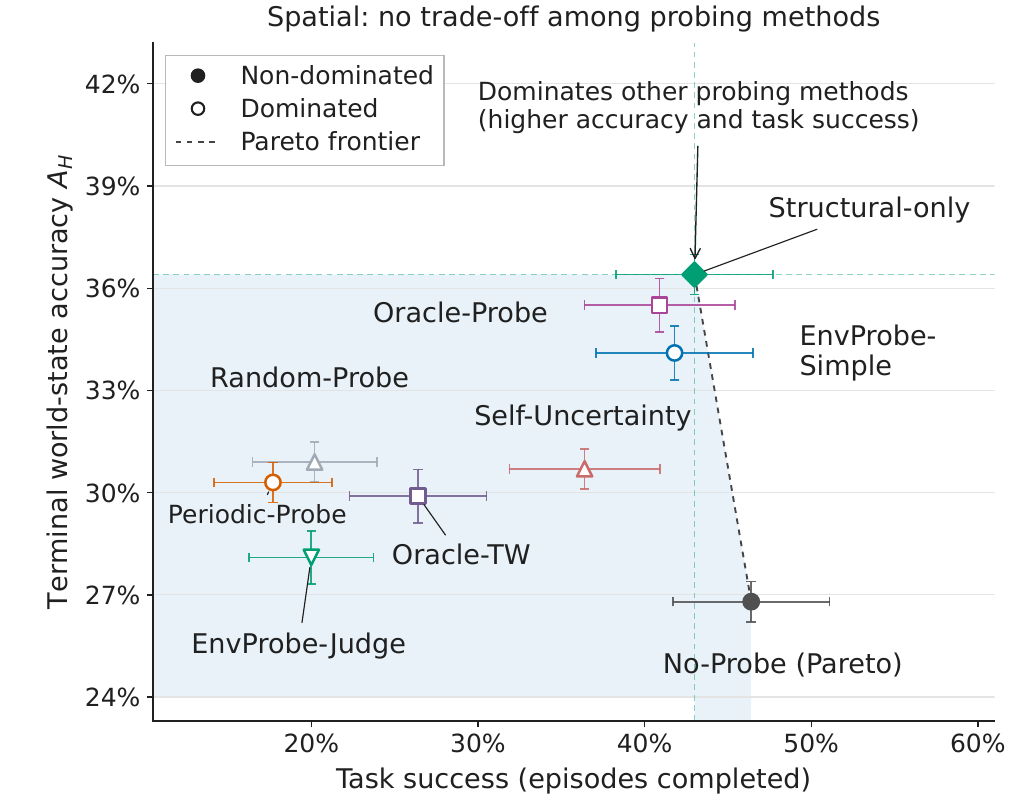}
		
		\caption{\textbf{Spatial belief--task Pareto frontier.} Unlike the procedural setting (Figure~\ref{fig:pareto_proc}), Structural-only dominates all baselines in both belief accuracy and task success. Shorter dependency chains reduce probing costs.}
		\label{fig:pareto_spatial}
		
	\end{figure}

	\section{Conclusion}
	\label{sec:conclusion}
	
	We studied when and why environment probing improves the reliability of world models in long-horizon language agents. We introduced \method{}, a framework that prioritizes belief verification using task criticality, dependency, staleness, and uncertainty, and established theoretical bounds on belief repair and its trade-off with task execution. Experiments across procedural and spatial environments show that targeted probing improves world-state accuracy, delays belief collapse, and exposes errors missed by self-reported uncertainty. Structural cues consistently provide the strongest guidance, while additional probing does not necessarily improve task success. These findings highlight the importance of task-aware belief verification and balancing information acquisition against the cost of interaction.
	
	\section*{Limitations}
	
	Our evaluation focuses on controlled environments that enable direct measurement of belief accuracy, which may not fully capture the complexity of real-world agent interactions. The effectiveness of environment probing may vary with observation quality, environmental dynamics, and model capabilities. Our theoretical results rely on assumptions about belief-repair gains and probe selection that may not hold universally. Further evaluation across diverse models, tasks, and deployment settings is needed to establish the generality and robustness of our findings.
	
	\bibliography{references}
	
	\clearpage
	\appendix

	\section{Theory}
	\label{sec:theory}
	
	We analyze how environment probing improves belief accuracy while reducing the interaction steps available for task execution. The analysis distinguishes belief types and characterizes the effects of probe selection quality, self-reported reliability, and interaction costs.
	
	\subsection{Belief-Side Gain}
	\label{sec:theory:belief}
	
	We analyze belief repair separately for each type $T\in\{\proc,\spat\}$. Let $\fieldset_T\subseteq\fieldset$ contain $n_T$ fields, and define $A_H^T=\frac{1}{n_T}\sum_{i\in\fieldset_T}\mathbf{1}\{b_H^i=g_H^i\}$ following Eq.~\eqref{eq:terminal_accuracy}. For a fixed trajectory prefix, the expected gain from probing $S\subseteq\fieldset_T$, relative to no probing, is
	\begin{equation}\label{eq:type_gain}
	\small
	G_T(S)=\mathbb{E}[A_H^T\mid\mathrm{do}(\mathrm{Probe}(S))]-\mathbb{E}[A_H^T\mid\mathrm{do}(\mathrm{Probe}(\emptyset))].\end{equation}
	Both expectations use the same continuation policy and remaining interaction horizon, so $G_T(S)$ includes the effects of probing on subsequent decisions.
	
	\begin{assumption}[Diminishing belief repair]
		\label{ass:submod}
		For each type $T$, the function $G_T$ in Eq.~\eqref{eq:type_gain} is normalized, monotone, and submodular. For all $S\subseteq R\subseteq\fieldset_T$ and $i\in\fieldset_T\setminus R$,
		\begin{equation}\label{eq:submodularity}\begin{aligned}
		G_T(\emptyset)&=0,\qquad G_T(S)\leq G_T(R),\\
		\Delta_T(i\mid S)&\geq\Delta_T(i\mid R),
		\end{aligned}\end{equation}
		where $\Delta_T(i\mid S)=G_T(S\cup\{i\})-G_T(S)$.
	\end{assumption}
	
	Assumption~\ref{ass:submod} captures diminishing returns in belief repair, consistent with submodular information gathering~\citep{nemhauser1978analysis,golovin2011adaptive,krause2014submodular}. It is a condition on the environment and continuation policy, rather than a property guaranteed by probing.
	
	\begin{definition}[Marginal selection quality]
		\label{def:gamma}
		For a fixed type $T$, let $i_\rho(S)$ denote the highest-scoring field among $\fieldset_T\setminus S$, following Eq.~\eqref{eq:probe_selection}. The score $\rho$ in Eq.~\eqref{eq:rho_def} has marginal selection quality $\gamma_T(\rho)\in[0,1]$ if, for every $S\subsetneq\fieldset_T$,
		\begin{equation}\label{eq:gamma_quality}\mathbb{E}[\Delta_T(i_\rho(S)\mid S)]\geq\gamma_T(\rho)\max_{j\in\fieldset_T\setminus S}\Delta_T(j\mid S).\end{equation}
	\end{definition}
	
	Definition~\ref{def:gamma} measures how closely score-based selection approximates the largest available marginal gain. Under Assumption~\ref{ass:submod}, $\gamma_T(\rho)=1$ corresponds to optimal marginal selection, while smaller values permit less effective targeting.
	
	\begin{lemma}[Targeted belief-repair bound]
		\label{lem:belief_bound}
		\label{thm:main_rev}
		Under Assumption~\ref{ass:submod} and Eq.~\eqref{eq:gamma_quality}, a greedy policy allocating $B_T$ probes to type $T$ satisfies
		\begin{equation}\label{eq:belief_bound}\mathbb{E}[G_T(S_{\rho,T})]\geq(1-e^{-\gamma_T(\rho)})\max_{\substack{S\subseteq\fieldset_T\\|S|\leq B_T}}G_T(S).\end{equation}
	\end{lemma}
	
	Lemma~\ref{lem:belief_bound} establishes that the belief gain defined in Eq.~\eqref{eq:type_gain} depends on selection quality, not merely the number of probes. Higher $\gamma_T(\rho)$ yields a stronger approximation guarantee under diminishing returns. The proof is provided in Appendix Section~\ref{app:proof_thm1}.
	
	\paragraph{Empirical selection quality.}
	\label{sec:method:tau_empirical}
	We assess selection quality through realized belief gains relative to Oracle-TW in Eq.~\eqref{eq:oracle_tw} under matched seeds. Comparing the full score in Eq.~\eqref{eq:rho_def} with the structural score in Eq.~\eqref{eq:structural_variant} evaluates the contribution of self-reported staleness and uncertainty.
	
	\begin{lemma}[Limited value of self-reported reliability]
		\label{lem:decomp_rev}
		Fix $t$ and $S\subsetneq\fieldset_{\spat}$. Suppose that, for every $i\in\fieldset_{\spat}\setminus S$,
		\begin{equation}\label{eq:self_report_eps}\begin{aligned}
		\bigl|&\mathbb{E}[\Delta_{\spat}(i\mid S)\mid c_i,d_i(t),s_i(t),u_i(t)]\\
		&-\mathbb{E}[\Delta_{\spat}(i\mid S)\mid c_i,d_i(t)]\bigr|\leq\varepsilon_{\spat}.
		\end{aligned}\end{equation}
		Assume the structural signals $c_i,d_i(t)$ and self-reported signals $s_i(t),u_i(t)$ are available jointly for all candidate fields. Then
		\begin{equation}\label{eq:self_report_bound}
		\small
		0\leq\sup_{\phi}\mathbb{E}[\Delta_{\spat}(\phi\mid S)]-\sup_{\psi}\mathbb{E}[\Delta_{\spat}(\psi\mid S)]\leq2\varepsilon_{\spat}.\end{equation}
	\end{lemma}
	
	Here, $\phi$ and $\psi$ range over all selection policies using the full signals in Eq.~\eqref{eq:rho_def} and the structural signals in Eq.~\eqref{eq:structural_variant}, respectively. Lemma~\ref{lem:decomp_rev} shows that self-reported reliability provides limited additional value when it weakly predicts marginal belief repair beyond task structure. The result concerns optimal selectors under the two information sets, rather than guaranteeing the performance of either fixed scoring rule. The proof is provided in Appendix Section~\ref{app:proof_lemma6}.
	
	\begin{proposition}[Uncertainty-only miscoverage]
		\label{lem:fano}
		Let $E_t=\{i\in\fieldset:b_t^i\neq g_t^i\}$ be the set of incorrect beliefs. For a threshold $\alpha\in[0,1]$, define $C_t=\{i\in E_t:u_i(t)\leq\alpha\}$ using Eq.~\eqref{eq:uncertainty}. If an uncertainty-only policy probes exclusively from $\{i:u_i(t)>\alpha\}$, then, for $|E_t|>0$,
		\begin{equation}\label{eq:miscoverage}\mathrm{Recall}_t\leq1-\frac{|C_t|}{|E_t|}.\end{equation}
	\end{proposition}
	
	Proposition~\ref{lem:fano} establishes that uncertainty-only probing cannot cover confidently incorrect beliefs excluded by its selection threshold. Its maximum error coverage is therefore limited by the fraction of incorrect beliefs assigned low uncertainty. The proof is provided in Appendix Section~\ref{app:proof_uncertainty}.

	\begin{proposition}[Non-adaptive allocation loss]
		\label{lem:nwf}
		For a fixed belief type $T$, let $q_i=G_T(\{i\})$ denote the single-field gain in Eq.~\eqref{eq:type_gain}, and suppose $\max_{i\in\fieldset_T}q_i>0$. The relative gain of uniform single-field probing satisfies
		\begin{equation}\label{eq:nonadaptive_loss}\mathrm{Eff}_{\mathrm{unif}}^T=\frac{\sum_{i\in\fieldset_T}q_i}{n_T\max_{i\in\fieldset_T}q_i}\leq1,\end{equation}
		with equality if and only if all single-field gains are equal.
	\end{proposition}
	
	Proposition~\ref{lem:nwf} quantifies the loss from ignoring heterogeneous belief-repair gains. Unlike the targeted selection rule in Eq.~\eqref{eq:probe_selection}, uniform probing assigns equal probability to all fields, regardless of their expected benefit. The proof is provided in Appendix Section~\ref{app:proof_nwf}.
	
	\subsection{Task-Side Displacement}
	\label{sec:theory:task}
	
	\begin{lemma}[Probe-action displacement]
		\label{lem:displacement}
		Let $P_\pi\leq B$ be the number of probes used by policy $\pi$ within horizon $H$. Suppose task success requires at least $K>0$ effective actions, and the conditional probability that any executed task action is effective, given the preceding history, is at most $\eta_\pi\in[0,1]$. Then
		\begin{equation}\label{eq:displacement}\Pr[\mathrm{success}(\pi)]\leq\min\left\{1,\frac{\eta_\pi\mathbb{E}[H-P_\pi]}{K}\right\}.\end{equation}
	\end{lemma}
	
	Lemma~\ref{lem:displacement} bounds task success by the expected number of available task actions. Since each probe consumes one interaction step, increasing $P_\pi$ reduces this bound unless probing also improves action effectiveness. The proof is provided in Appendix Section~\ref{app:displacement}.
	
	\begin{theorem}[Probe-action trade-off]
		\label{thm:tradeoff}
		Suppose policy $\pi$ allocates a fixed number $B_T$ of probes to each belief type $T\in\{\proc,\spat\}$, with $P_\pi=B_{\proc}+B_{\spat}\leq B$. Under Assumption~\ref{ass:submod} and Eq.~\eqref{eq:gamma_quality}, assume the policy satisfies the greedy selection guarantee in Lemma~\ref{lem:belief_bound}. Then its expected terminal belief gain satisfies
		\begin{equation}\label{eq:frontier_belief}
		\small
		\mathcal{B}(\pi)\geq\sum_{T\in\{\proc,\spat\}}\frac{n_T}{n}(1-e^{-\gamma_T(\rho)})\max_{\substack{S\subseteq\fieldset_T\\|S|\leq B_T}}G_T(S),\end{equation}
		where $\mathcal{B}(\pi)=\mathbb{E}[A_H(\pi)]-\mathbb{E}[A_H(\mathrm{NoProbe})]$, provided the type-specific gains in Eq.~\eqref{eq:type_gain} are evaluated against the same no-probe continuation. Under the effectiveness condition of Lemma~\ref{lem:displacement}, task success simultaneously satisfies
		\begin{equation}\label{eq:frontier_task}
		\small
		\Pr[\mathrm{success}(\pi)]\leq\min\left\{1,\frac{\eta_\pi(H-B_{\proc}-B_{\spat})}{K}\right\}.\end{equation}
	\end{theorem}
	
	Theorem~\ref{thm:tradeoff} combines the belief-repair guarantee in Eq.~\eqref{eq:belief_bound} with the action-displacement bound in Eq.~\eqref{eq:displacement}. Additional probes can increase the guaranteed belief gain while reducing the number of task-action opportunities. The resulting trade-off depends on selection quality, belief type, and action effectiveness. The proof is provided in Appendix Section~\ref{app:proof_tradeoff}.
	
	\section{Proofs}
	\label{app:proofs}
	
	We provide proofs of the theoretical results in Section~\ref{sec:theory}.
	
	\subsection{Proof of Lemma~\ref{lem:belief_bound}}
	\label{app:proof_thm1}
	
	\begin{proof}
		Fix a belief type $T$ and let $S_j$ denote the fields selected after $j$ probes, with $S_0=\emptyset$. Let $S_T^\star\in\arg\max_{S\subseteq\fieldset_T,|S|\leq B_T}G_T(S)$. By monotonicity and submodularity in Assumption~\ref{ass:submod},
		\begin{equation}\begin{aligned}
		G_T(S_T^\star)-G_T(S_j)
		&\leq G_T(S_j\cup S_T^\star)-G_T(S_j)\\
		&\leq\sum_{i\in S_T^\star\setminus S_j}\Delta_T(i\mid S_j)\\
		&\leq B_T\max_{i\in\fieldset_T\setminus S_j}\Delta_T(i\mid S_j).
		\end{aligned}\end{equation}
		
		Applying the marginal-quality condition in Eq.~\eqref{eq:gamma_quality} conditionally on $S_j$ gives
		\begin{equation}\begin{aligned}
		&\mathbb{E}[G_T(S_{j+1})-G_T(S_j)\mid S_j]\\
		&=\mathbb{E}[\Delta_T(i_\rho(S_j)\mid S_j)\mid S_j]\\
		&\geq\gamma_T(\rho)\max_{i\in\fieldset_T\setminus S_j}\Delta_T(i\mid S_j)\\
		&\geq\frac{\gamma_T(\rho)}{B_T}\left(G_T(S_T^\star)-G_T(S_j)\right).
		\end{aligned}\end{equation}
		
		Taking expectations and rearranging,
		\begin{equation}\begin{aligned}
		&G_T(S_T^\star)-\mathbb{E}[G_T(S_{j+1})]\\
		&\leq\left(1-\frac{\gamma_T(\rho)}{B_T}\right)\left(G_T(S_T^\star)-\mathbb{E}[G_T(S_j)]\right).
		\end{aligned}\end{equation}
		
		Iterating over $B_T$ probes and using $G_T(\emptyset)=0$ from Eq.~\eqref{eq:submodularity},
		\begin{equation}\begin{aligned}
		&G_T(S_T^\star)-\mathbb{E}[G_T(S_{B_T})]\\
		&\leq\left(1-\frac{\gamma_T(\rho)}{B_T}\right)^{B_T}G_T(S_T^\star)\\
		&\leq e^{-\gamma_T(\rho)}G_T(S_T^\star).
		\end{aligned}\end{equation}
		
		Therefore,
		\begin{equation}\begin{aligned}
		\mathbb{E}[G_T(S_{B_T})]
		&\geq\left(1-e^{-\gamma_T(\rho)}\right)G_T(S_T^\star)\\
		&=\left(1-e^{-\gamma_T(\rho)}\right)\max_{\substack{S\subseteq\fieldset_T\\|S|\leq B_T}}G_T(S),
		\end{aligned}\end{equation}
		which establishes Eq.~\eqref{eq:belief_bound}.
	\end{proof}
	
	\subsection{Proof of Lemma~\ref{lem:decomp_rev}}
	\label{app:proof_lemma6}
	
	\begin{proof}
		Fix $t$ and $S\subsetneq\fieldset_{\spat}$. Let $\phi^\star$ and $\psi^\star$ be optimal selectors using the full signals in Eq.~\eqref{eq:rho_def} and the structural signals in Eq.~\eqref{eq:structural_variant}, respectively. Denote their selected fields by $i_\phi$ and $i_\psi$.
		
		By the conditional gain bound in Eq.~\eqref{eq:self_report_eps}, the expected marginal gain of any candidate changes by at most $\varepsilon_{\spat}$ when self-reported signals are included. Since $\psi^\star$ maximizes expected gain using structural information, we obtain
		\begin{equation}\begin{aligned}
		&\mathbb{E}[\Delta_{\spat}(i_\phi\mid S)]\\
		&\leq \mathbb{E}[\Delta_{\spat}(i_\phi\mid S)\mid c_{i_\phi},d_{i_\phi}(t)]+\varepsilon_{\spat}\\
		&\leq \mathbb{E}[\Delta_{\spat}(i_\psi\mid S)\mid c_{i_\psi},d_{i_\psi}(t)]+\varepsilon_{\spat}\\
		&\leq \mathbb{E}[\Delta_{\spat}(i_\psi\mid S)]+2\varepsilon_{\spat}.
		\end{aligned}\end{equation}
		
		Taking the supremum over admissible selectors yields
		\begin{equation}\begin{aligned}
		0
		&\leq \sup_{\phi}\mathbb{E}[\Delta_{\spat}(\phi\mid S)]-\sup_{\psi}\mathbb{E}[\Delta_{\spat}(\psi\mid S)]\\
		&\leq 2\varepsilon_{\spat},
		\end{aligned}\end{equation}
		which establishes Eq.~\eqref{eq:self_report_bound}.
	\end{proof}
	
	\subsection{Proof of Proposition~\ref{lem:fano}}
	\label{app:proof_uncertainty}
	
	\begin{proof}
		By Eq.~\eqref{eq:uncertainty}, an uncertainty-only policy with threshold $\alpha$ selects fields satisfying $u_i(t)>\alpha$. Hence, every incorrectly believed field in $C_t$, as defined in Proposition~\ref{lem:fano}, is excluded from the eligible set. For $|E_t|>0$, the maximum fraction of incorrect fields covered by the eligible set satisfies
		\begin{equation}\begin{aligned}
		\mathrm{Recall}_t
		&\leq\frac{|\{i\in E_t:u_i(t)>\alpha\}|}{|E_t|}\\
		&=\frac{|E_t\setminus C_t|}{|E_t|}\\
		&=1-\frac{|C_t|}{|E_t|}.
		\end{aligned}\end{equation}
		This establishes Eq.~\eqref{eq:miscoverage}.
	\end{proof}
	
	\begin{table*}[!ht]
		\centering
		\caption{\textbf{Budget-normalized useful-probe rate $R_B$.} Green highlights the highest value in each environment.}
		\label{tab:g4_norm}
		\small
		\setlength{\tabcolsep}{4pt}
		\resizebox{0.8\linewidth}{!}{
			\begin{tabular}{lrrrrr}
				\toprule
				\textbf{Environment} & \textbf{\method-Simple} & \textbf{Random-Probe} & \textbf{Periodic-Probe} & \textbf{Self-Uncertainty} & \textbf{Oracle-Probe} \\
				\midrule
				GraphNavWorld & 0.91 & 0.49 & 0.74 & 0.46 & \cellcolor{bestgreen}\textbf{0.97} \\
				ObjectStateWorld & 0.31 & 0.29 & 0.67 & \cellcolor{bestgreen}\textbf{0.75} & 0.20 \\
				Spatial pool & 0.61 & 0.39 & \cellcolor{bestgreen}\textbf{0.71} & 0.61 & 0.58 \\
				ToolDAGWorld (diagnostic) & 0.00 & 0.19 & 0.14 & \cellcolor{bestgreen}\textbf{0.70} & 0.00 \\
				\bottomrule
			\end{tabular}
		}
	\end{table*}
	
	\subsection{Proof of Proposition~\ref{lem:nwf}}
	\label{app:proof_nwf}
	
	\begin{proof}
		Fix a belief type $T$ and let $q_i=G_T(\{i\})$, as defined in Proposition~\ref{lem:nwf}. Under uniform single-field probing, each field in $\fieldset_T$ is selected with probability $1/n_T$. By Eq.~\eqref{eq:type_gain}, the expected gain is
		\begin{equation}\begin{aligned}
		\mathbb{E}[G_T(\{i\})]
		&=\sum_{i\in\fieldset_T}\frac{1}{n_T}G_T(\{i\})\\
		&=\frac{1}{n_T}\sum_{i\in\fieldset_T}q_i\\
		&\leq\max_{i\in\fieldset_T}q_i.
		\end{aligned}\end{equation}
		
		Since $\max_{i\in\fieldset_T}q_i>0$, dividing both sides by this maximum gives
		\begin{equation}\begin{aligned}
		\mathrm{Eff}_{\mathrm{unif}}^T
		&=\frac{\mathbb{E}[G_T(\{i\})]}{\max_{j\in\fieldset_T}q_j}\\
		&=\frac{\sum_{i\in\fieldset_T}q_i}{n_T\max_{j\in\fieldset_T}q_j}\\
		&\leq1.
		\end{aligned}\end{equation}
		
		Equality holds if and only if $q_i=\max_{j\in\fieldset_T}q_j$ for every $i\in\fieldset_T$. Otherwise, the inequality is strict, establishing Eq.~\eqref{eq:nonadaptive_loss}.
	\end{proof}
	
	\subsection{Proof of Lemma~\ref{lem:displacement}}
	\label{app:displacement}
	
	\begin{proof}
		Let $M_\pi$ denote the number of effective task actions executed by policy $\pi$ within horizon $H$. Since each probe consumes one interaction step, the number of executed task actions is at most $H-P_\pi$.
		
		By the conditional effectiveness assumption in Lemma~\ref{lem:displacement}, the expected number of effective task actions satisfies
		\begin{equation}\begin{aligned}
		\mathbb{E}[M_\pi]
		&=\mathbb{E}\left[\sum_{t=1}^{H}\mathbf{1}\{\text{effective task action at }t\}\right]\\
		&\leq\eta_\pi\mathbb{E}[H-P_\pi].
		\end{aligned}\end{equation}
		
		Task success requires at least $K$ effective actions. Applying Markov's inequality therefore gives
		\begin{equation}\begin{aligned}
		\Pr[\mathrm{success}(\pi)]
		&\leq\Pr[M_\pi\geq K]\\
		&\leq\frac{\mathbb{E}[M_\pi]}{K}\\
		&\leq\frac{\eta_\pi\mathbb{E}[H-P_\pi]}{K}.
		\end{aligned}\end{equation}
		
		Combining this inequality with $\Pr[\mathrm{success}(\pi)]\leq1$ yields
		\begin{equation}\begin{aligned}
		\Pr[\mathrm{success}(\pi)]
		&\leq\min\left\{1,\frac{\eta_\pi\mathbb{E}[H-P_\pi]}{K}\right\},
		\end{aligned}\end{equation}
		which establishes Eq.~\eqref{eq:displacement}.
	\end{proof}
	
	\subsection{Proof of Theorem~\ref{thm:tradeoff}}
	\label{app:proof_tradeoff}
	
	\begin{proof}
		Fix the probe allocation $B_{\proc},B_{\spat}$ in Theorem~\ref{thm:tradeoff}, and let $S_{\rho,T}$ denote the fields selected by the score-based policy for each $T\in\{\proc,\spat\}$.
		
		By Eq.~\eqref{eq:terminal_accuracy}, terminal belief accuracy decomposes as
		\begin{equation}\begin{aligned}
		A_H
		&=\frac{1}{n}\sum_{i\in\fieldset}\mathbf{1}\{b_H^i=g_H^i\}\\
		&=\sum_{T\in\{\proc,\spat\}}\frac{1}{n}\sum_{i\in\fieldset_T}\mathbf{1}\{b_H^i=g_H^i\}\\
		&=\sum_{T\in\{\proc,\spat\}}\frac{n_T}{n}A_H^T.
		\end{aligned}\end{equation}
		
		Using Eq.~\eqref{eq:type_gain}, the common no-probe continuation, and the type-specific gain decomposition assumed in Theorem~\ref{thm:tradeoff}, we obtain
		\begin{equation}\begin{aligned}
		&\mathcal{B}(\pi)\\
		&=\mathbb{E}[A_H(\pi)]-\mathbb{E}[A_H(\mathrm{NoProbe})]\\
		&=\sum_{T\in\{\proc,\spat\}}\frac{n_T}{n}\left(\mathbb{E}[A_H^T(\pi)]-\mathbb{E}[A_H^T(\mathrm{NoProbe})]\right)\\
		&=\sum_{T\in\{\proc,\spat\}}\frac{n_T}{n}\mathbb{E}[G_T(S_{\rho,T})]\\
		&\geq\sum_{T\in\{\proc,\spat\}}\frac{n_T}{n}(1-e^{-\gamma_T(\rho)})\max_{\substack{S\subseteq\fieldset_T\\|S|\leq B_T}}G_T(S),
		\end{aligned}\end{equation}
		where the inequality follows from Eq.~\eqref{eq:belief_bound}. This proves Eq.~\eqref{eq:frontier_belief}.
		
		For task success, the fixed probe allocation gives $P_\pi=B_{\proc}+B_{\spat}$. Applying Eq.~\eqref{eq:displacement},
		\begin{equation}\begin{aligned}
		\Pr[\mathrm{success}(\pi)]
		&\leq\min\left\{1,\frac{\eta_\pi\mathbb{E}[H-P_\pi]}{K}\right\}\\
		&=\min\left\{1,\frac{\eta_\pi(H-\mathbb{E}[P_\pi])}{K}\right\}\\
		&=\min\left\{1,\frac{\eta_\pi(H-B_{\proc}-B_{\spat})}{K}\right\},
		\end{aligned}\end{equation}
		which proves Eq.~\eqref{eq:frontier_task}.
		
		Finally, monotonicity in Eq.~\eqref{eq:submodularity} implies
		\begin{equation}\begin{aligned}
		\max_{\substack{S\subseteq\fieldset_T\\|S|\leq B_T}}G_T(S)
		&\leq\max_{\substack{S\subseteq\fieldset_T\\|S|\leq B_T+1}}G_T(S),
		\end{aligned}\end{equation}
		whereas, for fixed $\eta_\pi$,
		\begin{equation}\begin{aligned}
		&\min\left\{1,\frac{\eta_\pi(H-B_{\proc}-B_{\spat}-1)}{K}\right\}\\
		&\leq\min\left\{1,\frac{\eta_\pi(H-B_{\proc}-B_{\spat})}{K}\right\}.
		\end{aligned}\end{equation}
		
		Thus, increasing the probe allocation weakly increases the optimal belief-repair gain while weakly decreasing the task-success upper bound, establishing the claimed probe-action trade-off.
	\end{proof}
	


	
	
	
	\section{Useful-Probe Rate Diagnostics}
	
	We evaluate probe efficiency using the useful-probe rate $\mathrm{UPR}$, defined as the fraction of executed probes that correct an incorrect belief, and the budget-normalized rate $R_B=\#\mathrm{useful}/B$, which accounts for differences in probe frequency. Both metrics complement the analysis in Section~\ref{sec:results:secondary}.
	
	\begin{table}[!ht]
		\centering
		\caption{\textbf{Useful-probe rate across environments.} ToolDAGWorld results are diagnostic due to uninstantiated procedural fields.}
		\label{tab:g4_main}
		\small
		\setlength{\tabcolsep}{5pt}
		\resizebox{\columnwidth}{!}{
			\begin{tabular}{lrr}
				\toprule
				\textbf{Environment} & \textbf{\method-Simple} & \textbf{Random-Probe} \\
				\midrule
				Spatial pool & \cellcolor{bestgreen}\textbf{0.735} & 0.550 \\
				Combined & \cellcolor{bestgreen}\textbf{0.477} & 0.460 \\
				ToolDAGWorld (diagnostic) & 0.000 & \cellcolor{bestgreen}\textbf{0.294} \\
				\bottomrule
			\end{tabular}
		}
	\end{table}
	
	Table~\ref{tab:g4_main} shows that \method-Simple achieves a higher useful-probe rate than Random-Probe on spatial tasks (73.5\% versus 55.0\%). However, raw rates depend on probe frequency, and procedural results are affected by scorer limitations.

	Table~\ref{tab:g4_norm} shows that budget normalization reveals substantial variation across environments. \method-Simple approaches Oracle-Probe on GraphNavWorld, while Self-Uncertainty performs best on ObjectStateWorld and Periodic-Probe leads on the spatial pool. Procedural results remain diagnostic because of scorer limitations.

\end{document}